\documentclass{article}

\usepackage{PRIMEarxiv}

\usepackage[utf8]{inputenc} 
\usepackage[T1]{fontenc}    
\usepackage{hyperref}       
\usepackage{url}            
\usepackage{booktabs}       
\usepackage{amsfonts}       
\usepackage{nicefrac}       
\usepackage{microtype}      
\usepackage{lipsum}
\usepackage{fancyhdr}       
\usepackage{graphicx}       
\graphicspath{{media/}}     
\usepackage{caption}
\usepackage{placeins}
\usepackage{float}

\title{Chronic Diseases Prediction Using ML
}

\author{
 Sri Varsha Mulakala , G.Neeharika, P.Vinay Kumar, A.Bhargava Kiran 
}

\begin{document}
\maketitle

\begin{abstract}
The recent increase in morbidity is primarily due to chronic diseases including Diabetes, Heart disease, Lung cancer, and brain tumours. The results for patients can be improved, and the financial burden on the healthcare system can be lessened, through the early detection and prevention of certain disorders. In this study, we built a machine-learning model for predicting the existence of numerous diseases utilising datasets from various sources, including Kaggle, Dataworld, and the UCI repository, that are relevant to each of the diseases we intended to predict.

Following the acquisition of the datasets, we used feature engineering to extract pertinent features from the information, after which the model was trained on a training set and improved using a validation set. A test set was then used to assess the correctness of the final model. We provide an easy-to-use interface where users may enter the parameters for the selected ailment. Once the right model has been run, it will indicate whether the user has a certain ailment and offer suggestions for how to treat or prevent it.

\end{abstract}


\section{Introduction}
Data is a valuable resource in the digital age, with vast amounts of data being generated in every industry. All patient-related data is contained in healthcare data. A general architecture for disease prediction in healthcare has been published. In some current models, he focuses on only one disease in each analysis. For example, no mechanism can simultaneously analyze multiple diseases, such as cancer once, diabetes, and brain tumour once. We concentrate on giving consumers rapid and precise disease predictions based on their entered symptoms and anticipated ailments.

Therefore, using Streamlit, we present a system that can be used to predict various diseases. This system analyses the status of diabetes, heart cancer, lung cancer, and brain tumours. More diseases may be added later. Build a chronic disease prediction system using convolutional neural networks and machine learning techniques. After training the model with the provided data, save the behaviour of the model.

The creation of a system that can reliably forecast a variety of diseases is the major objective of this project. Thanks to this project, users do not need to visit various websites, which also saves them time. Diseases predicted early can extend life expectancy and save you from financial troubles. Therefore, various machine learning and CNN algorithms were used here to achieve maximum accuracy. Our results demonstrate the effectiveness of these techniques in predicting chronic disease risk and improving clinical outcomes.  
\begin{figure}[h]
    \centering
    \includegraphics[width=0.3\textwidth]{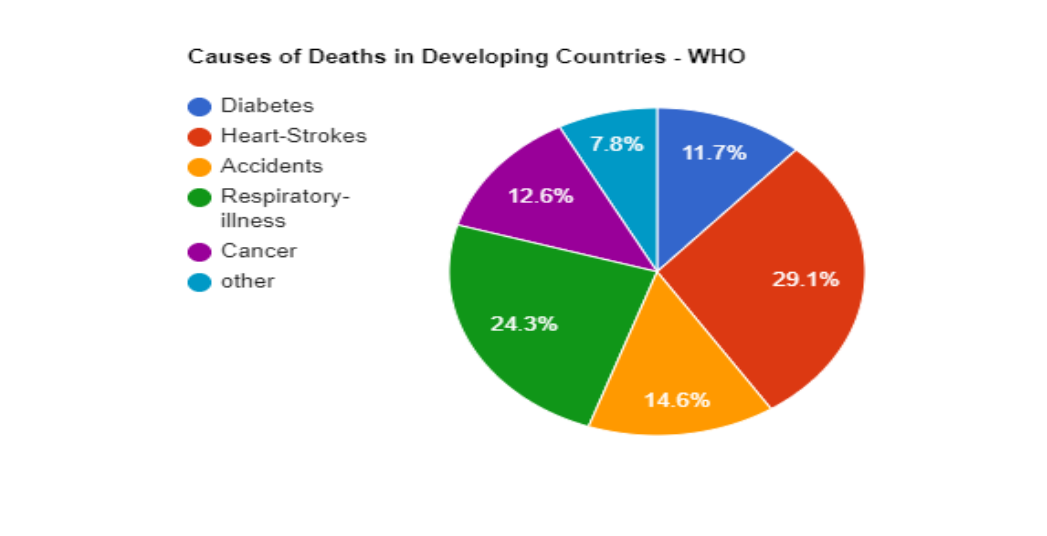} 
    \caption{Pie-Chart of Various Chronic Diseases}
    \label{fig:chronic_diseases}
\end{figure}

\section{ Literature Review}
Yasuda et al. categorises the several kinds of datasets that might be produced in order to ascertain if a person has diabetes or not. The hospital warehouse, which contains information from 200 instances with nine attributes, is used to generate the data collection for diabetic patients. These dataset examples are from the categories of urine and blood testing. WEKA can be utilised to classify the data in this investigation because of its great performance on tiny datasets. The results are compared after the data has been assessed using a 10-fold cross-validation approach. The naïve Bayes, J48, REP Tree, and Random Tree are used. The findings revealed that J48, with an accuracy of 60.2\%, outperforms the competition.

By exploring and examining the patterns emerging in the data via classification analysis utilising Decision Tree and Nave Bayes algorithms, Aiswarya et al. seek to identify ways to diagnose diabetes. The goal of the research is to provide a quicker and more effective way to diagnose the condition, which will aid in the patients' prompt recovery. The J48 algorithm provides an accuracy rate of 74.8\% whereas the naive Bayes provides an accuracy of 79.5\% by employing a 70:30 split, according to the study's findings using the PIMA dataset and cross-validation methodology.

The decision tree classifier, KNN, SVM, Random classifier, and Heart Disease Prediction System-HDPS (O.E. Taylor) is depicted and it is equated with delivered an exact result. As a result, when compared to previous methods, the decision tree machine learning technology was able to anticipate occurrences with an improved accuracy of 98.85\%.

Heart disease incidence forecasting and analysis ChalaBeyene and colleagues suggested using data mining techniques. To swiftly and automatically identify cardiac disease, the main objective is to anticipate its onset. In a healthcare institution where professionals lack current skills and knowledge, the suggested strategy is crucial. Numerous medical criteria, such as blood sugar levels, heart rate, age, and sex, are utilised to assess whether or not a person has a cardiac condition. Analyses of the dataset are carried out using WEKA software. This results in a precision of 81.34\%. 

The authors of the article outline a useful technique for identifying and classifying lung cancer in CT scan pictures. They employed decision trees, random forests, support vector machines, naive Bayes, k-nearest neighbours, stochastic gradient descent, and multi-layer perceptrons among the seven classification models. For the training and testing of these classifiers, a dataset of 15,750 clinical data containing 6910 benign and 8840 malignant lung cancer-related images was used. With an accuracy rating of 88.55\%, the multi-layer perceptron classifier surpassed the other classifiers in the results. 

The authors of the study forecast lung cancer using a number of techniques, including a neural network, radial basis function network, support vector machine, logistic regression, random forest, J48, naive Bayes, and K-nearest neighbours. When used with data pertaining to lung cancer, they showed that the radial basis function network had a superior accuracy of 81.25\%. The early diagnosis of lung cancer by assessing the efficacy of classification algorithms is likewise the main objective [35]. The authors employed naive Bayes, support vector machines, decision trees, logistic regression, and other classification algorithms. In comparison to the lung cancer dataset from the UCI, where logistic regression had a greater accuracy of 96.9\%, support vector machine (SVM) obtained a higher accuracy of 99.2\%.

To discriminate between brain regions with and without tumours, they employed the Fuzzy segmentation approach (FCM) in their research. In order to obtain wavelet features, a multilayer discrete wavelet transform was also applied. (DWT). To precisely diagnose brain cancers, deep neural networks (DNNs) were used. In comparison to the KNN classifier, linear discriminant analysis (LDA), and sequential minimum optimisation (SMO) approaches, this strategy's efficacy was assessed. The accuracy percentage was 96.97\% in the DNN-based brain tumour categorization study. However, both the efficiency and the complexity were extremely poor.

Indian Pives, the Kennedy Space Center, and the University of Pavia provided the data for their research. The CNN technique was used, and the accuracy obtained was 88.75\%. The material was obtained from the Cancer Imaging Archive. (TCIA). SVM, RF, LOG, MLP, and PCA classifiers were used, and KNN was the technique. The accuracy percentage for the proposed method was 83\%. Cheng's Fig Share provided the details. Using the CNN (Convolutional Neural Network) method, an accuracy of 84.19\% was achieved.
  
\section{Methodology}
\label{sec:headings}
\subsection{Machine Learning:}

In the field of artificial intelligence, machine learning is referred to as a subset that focuses on creating algorithms that let computers learn on their own from data and experience. When new data is acquired, a machine learning system predicts its outcome using a predictive model that it has built using historical data. The quantity of data affects how accurate the anticipated result is. Richer models that more precisely forecast the outcome can be created with the use of massive amounts of data.
Let's say you have a challenging situation where you need to forecast the future. You may simply input your data into a common algorithm and utilise those methods to have the machine create logic according to the data to anticipate the version without having to write any code. The way we approach challenges has changed as a result of machine learning. The machine learning algorithm's operation is depicted in the following block diagram. 

\begin{figure}[h]
    \centering
    \includegraphics[width=0.3\textwidth]{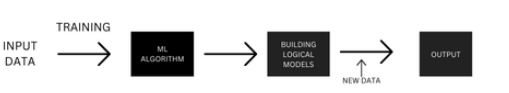} 
    \caption{ML Algorithm Work Flow}
    \label{fig:ml_workflow}
\end{figure}

One method of machine learning is supervised learning, in which you give a machine learning system labelled sample data to train on and anticipate an output from.

In order to create a model, comprehend the data set, and get additional knowledge about each data, the system employs the labelled data. Test the model after training and processing to check if it accurately predicts outputs by supplying sample data.

Mapping input data to output data is the objective of supervised learning. Learning that is supervised is similar to learning that takes place while students are being watched over by a teacher. Spam filtering is one application of supervised learning. 

\subsubsection{The Random Forest Classifier}
As the name suggests, a random forest consists of a large number of individual decision trees acting as an ensemble. Each tree in the random forest spits out class predictions, and the class with the most votes becomes the model's prediction (see figure below). 
\begin{figure}[h]
    \centering
    \includegraphics[width=0.3\textwidth]{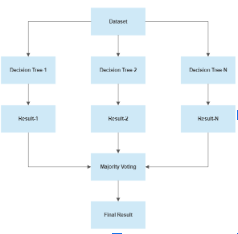} 
    \caption{Flowchart of Random Forest Model}
    \label{fig:random_forest}
\end{figure}
The wisdom of crowds is the fundamental tenet of Random Forests, a straightforward but effective method. In data science, random forest models perform better than each of its component models alone because they are composed of several largely uncorrelated models (trees). It is crucial for models to have little correlation. Since low-correlation assets (like stocks and bonds) tend to join together to form portfolios that are greater than the sum of their parts, an uncorrelated model can generate aggregate forecasts that are more accurate than individual forecasts. The trees prevent each other from making errors, unless they consistently make the same error in the same direction, which is what causes this pleasing result. 
While some trees might be mistaken, the majority of them are likely right, which allows the trees as a whole to move in the appropriate direction. Consequently, the following are necessary for good random forest performance:
These functions must have genuine signals in order for the models they are used in to outperform random guessing. Individual tree predictions (and errors) ought to have a weak correlation with one another. 

\subsection{Deep Learning:}
A subset of machine learning called deep learning employs artificial neural networks to offer computers the ability to learn from data and make judgements. Deep learning has contributed to some of the most astounding developments in recent years in artificial intelligence, including speech and picture recognition, natural language processing, and game-changing innovations.

The brain of deep learning is an artificial neural network that is modelled after the structure and operation of the human brain. These network layers' interconnected nodes, known as neurons, are employed to process and send information. Each neuron takes data from a different neuron, processes it mathematically, and then passes it to the layer of neurons below it.
 
Deep learning models may be trained on enormous labelled data sets, which enables them to find patterns and traits related to certain tasks. To learn to recognise various objects and patterns in photographs, for instance, deep learning models for image recognition can be trained on millions of labelled images. For applications like security systems, self-driving automobiles, and medical diagnostics, this model can be used to categorise fresh photographs and offer important information. Deep learning can automatically learn complicated features and representations from data without the need for explicit feature engineering, which is one of its key benefits. Therefore, complex, high-dimensional applications like speech and picture recognition are particularly well suited for deep learning. 

Deep learning models, however, might call for a lot of labelled training data and are computationally demanding. Additionally problematic are interpretability and explainability, as it can be challenging to comprehend how deep-learning models arrive at conclusions. Deep learning is a rapidly growing discipline with a lot of potentials to progress artificial intelligence and solve important issues, despite these challenges.

\subsubsection{Convolution Neural Networks:}
Convolutional neural networks (CNN) are the most prevalent type of deep neural network used for image recognition. CNNs can also be used for object recognition and other computer vision applications. With the use of multi-level convolutional filters, CNNs are made to automatically recognise and extract features from input images.
The CNN design includes pooling layers, convolutional layers, and fully linked layers. To extract features like edges, corners, and other patterns, input images are passed through a number of learnable filters (kernels) in convolutional layers. Convolutional layer feature maps formed by pooling layers are made less dimensional. The final classification of the images into one or more categories is based on the features collected from the fully linked layer.

Comparing convolutional neural networks to more established image identification techniques, there are various benefits. First, instead of having to manually create features, it can automatically learn and extract features from input images. The ability to understand complicated structures and patterns within images is made possible by the fact that it is very scalable and can be trained on huge datasets of annotated images.

And lastly, it can be used to solve a variety of computer vision issues such as semantic segmentation, object recognition, and image classification. ResNet, GoogleNet, VGGNet, and AlexNet are the most often used CNN designs. For a number of image recognition benchmarks, including the ImageNet Large Scale Visual Recognition Challenge, these models produced state-of-the-art results (ILSVRC). CNNs can also be utilised for a wide range of tasks, such as facial recognition, medical diagnosis, self-driving automobiles, and more.

The below figure shows the basic architecture of a convolution neural network.
\begin{figure}[h]
    \centering
    \includegraphics[width=0.3\textwidth]{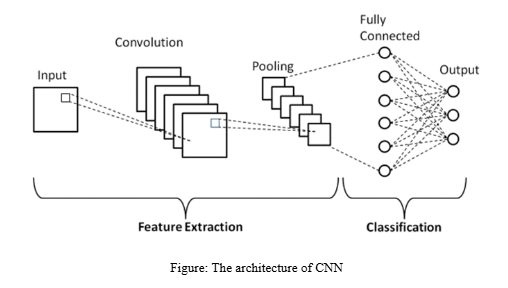} 
    \caption{The Architecture of CNN}
    \label{fig:cnn_architecture}
\end{figure}

\section{Proposed Method}
In our proposed system we have followed the subsequent steps: Data acquisition, Data pre-processing, Model, Training and Testing.
\paragraph{Diabetes:}
We have preprocessed the dataset which was taken from the Pima Indian diabetes dataset, and after certain intervals with the help of a heatmap, we did feature selection which contributes a major impact on the prediction of diabetes. The model we adopted was random forest, Ada boost, bagging, XGB classifier, Logistic Regression and a few models, So out of all models, Random forest gave better accuracy with 91.0 which we eventually stick with.
\begin{figure}[h]
\centering
    \includegraphics[width=0.3\textwidth]{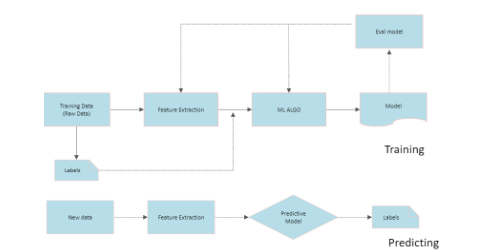} 
    \caption{Flow chart of Diabetes prediction Algo}
    \label{fig:flow_chart}
\end{figure}

\paragraph{Heart Disease:}
In order to forecast heart disease, we used the Public health dataset from Kaggle. During feature selection, we looked at a number of features and decided to include them in the specification. So, after that, we first used SVM, Gradient boosting, XGB classifier, Decision tree, Random forest, and a few models. In comparison to the other models we looked at, Random forest provided the best accuracy with 98.5.

\begin{figure}
    \centering
    \includegraphics[width=0.5\textwidth]{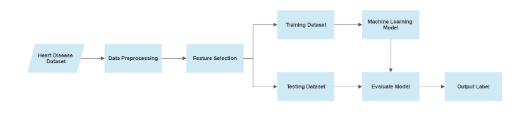}
    \caption{Flow chart of heart disease algorithm}
    \label{fig:flow_heart}
\end{figure}

\paragraph{Lungs Cancer:}
After obtaining the images from the Histopathological Imaging Dataset, we subjected them to preprocessing, including scaling and one hot encoding, which converts categorical data into numerical data. Using train-test splits, the model was subsequently trained and validated.

We have used a sequential model using the components listed below.
\begin{itemize}
    \item A convolutional kernel is built with three convolutional layers and is wound with input layers to help construct a tensor of outputs.

    \item The highest value for each input channel over an input window is utilised by MaxPooling Layers to downsample the input along its spatial dimensions. This is done after the convolutional layer.

    \item Before the final layer, we've added a Dropout layer to exclude any possibility of overfitting.

    \item For each of the three classes, the output layer—the bottom layer—generates soft probabilities.
\end{itemize}

\paragraph{Brain tumour:}
After obtaining the photos from the Br35H Brain Tumour Dataset, we preprocessed them using the label binarizer, a SciKit Learn class that takes categorical data as input and produces a Numpy array. We used the Data Augmentation technique since our data collection was so small that it was necessary to "raise" the size of the training set. Each training image in the batch is transformed (including by random rotation, scaling, etc.). The initial batch is then replaced with the newly created, randomly changed batch. After that, the recently converted batch is divided into a training dataset and a testing dataset.

Now, the upgraded set of pictures will include the VGG-16 model. The VGG-16 model can be used to evaluate the spectrum of tumour severity. There are 16 convolutional layers in the Visual Geometric Group-16 or VGG-16. With fixed image sizes of 224 224 (RGB), it has three channels: red, green, and blue. The VGG-16 has the following features: softmax, fully coneccted+Relu, max pooling, and convolutional+Relu. Before being divided into tumour-type and non-tumour-type categories, the image in this VGG-16 will go through each of these steps.

\section{Comparative Result and Analysis}
\label{sec:headings}
\subsection{Data collection}
\paragraph{Diabetes}
The Pima Indians Diabetes Dataset was used in this analysis and was downloaded from Kaggle. The datasets include one target variable, Outcome, together with a number of medical predictor variables. The patient's BMI, insulin level, age, number of previous pregnancies, and other factors are predictor variables.

After preprocessing the features selected are: 

\begin{table}[h]
    \centering
    \begin{tabular}{|c|c|c|}
        \hline
        SI.No. & Feature & Feature Description  \\
        \hline
        1.   & Pregnancies  & Number of times pregnant    \\
        \hline
        2.   & Glucose   &  Plasma glucose concentration a 2 hours in an oral glucose tolerance test  \\
        \hline
        3.  & Insulin   & 2-Hour serum insulin (mu U/ml)   \\
        \hline
        4.   & BMI  & Body mass index   \\
        \hline
        5.   & Age   &  Age(years) \\
        \hline
       
    \end{tabular}
    
    \label{tab:example}
\end{table}

\paragraph{Heart Disease}
The dataset utilised in this instance is the Public Health Dataset from Kaggle. Although this dataset has 76 attributes total, including the projected attributes, all published studies only use a portion of these 14. The patient has a cardiac illness, as indicated by the "Target" field. It is an integer where 0 indicates no disease and 1 indicates a disease.

After preprocessing the features selected are: 

\begin{table}[h]
    \centering
    \begin{tabular}{|c|c|c|}
        \hline
        SI.No. & Feature & Feature Description  \\
        \hline
        1.   & Age  & Age in years    \\
        \hline
        2.   & Sex   &  (1 = male; 0 = female)  \\
        \hline
        3.  & cp   & chest pain type   \\
        \hline
        4.   & trestbps  & resting blood pressure (in mm Hg on admission to the hospital)   \\
        \hline
        5.   & restecg  &  resting electrocardiographic results \\
        \hline
         6.   & thalach  & maximum heart rate achieved    \\
        \hline
        7.   & exang  & exercise-induced angina (1 = yes; 0 = no \\
        \hline
        8.  & oldpeak   & ST depression induced by exercise relative to rest   \\
        \hline
        9.   & slope  & the slope of the peak exercise ST segment    \\
        \hline
        10.   & Ca   &  number of major vessels (0-3) coloured by fluoroscopy \\
        \hline
        11.   & thal   &  1 = normal; 2 = fixed defect; 3 = reversible defect \\
        \hline
       
    \end{tabular}
    
    \label{tab:example}
\end{table}

\paragraph{Lung Cancer}
Here, the Histopathological Image Dataset from Kaggle is used as the dataset. There are 5 classifications and 25,000 histopathology pictures in this collection. Each image is 768 by 768 pixels in size and is stored as a JPEG file. 15000 photos of lung tissue were included in the images, which were created from original samples obtained from HIPAA-compliant validated sources (5000 images of benign lung tissue, 5000 images of lung adenocarcinoma, and 5000 images of lung squamous cell carcinoma).
The dataset we consider is divided into three classes, each containing 5000 images:
Benign lung tissue:  An unneeded abnormal growth of tissue that is not malignant is referred to as a benign lung tumour. Benign lung tumours can develop from a variety of different lung structures.
Lung adenocarcinoma: Non-small cell lung cancer of the lung called lung adenocarcinoma makes for around 40\% of all lung cancer cases. Promising novel therapeutic modalities are being explored in addition to those already in use.
Squamous cell carcinoma of the lung: Lung squamous cell carcinoma is a form of lung cancer. It happens when aberrant lung cells proliferate uncontrollably and develop into a tumour. Eventually, cancerous cells can "metastasize" to the liver, bones, brain, and adrenal glands, among other organs.

\paragraph{Brain Tumor}
This article's dataset, the  Br35H Brain Tumor Dataset, was obtained via Kaggle. 3060 brain MRI images are contained in the dataset's three files, yes, no, and pred. Brain tumours are difficult. The growth and distribution of brain tumours exhibit numerous anomalies. It is quite challenging to completely comprehend the nature of tumours as a result. Additionally, a qualified neurosurgeon is needed for MRI analysis. 
It is frequently highly challenging and time-consuming to compile reports from MRIs in poor nations because there aren't enough trained doctors and experts in the field. Therefore, this issue can be resolved by a cloud-based automated method.

\subsection{Data preprocessing}

Preparing raw data to be used by machine learning models is known as data preparation. This is a crucial first step in developing a machine-learning model. 
Not always will you find clear, well-formatted data while developing machine learning applications. Additionally, it is essential to always store the data in a neat and formatted manner before working with it. We use a data preprocessing task to do this.
Characteristics:
\begin{itemize}
    \item Getting the dataset
    \item Importing libraries
    \item Importing datasets
    \item Finding Missing Data
    \item Encoding Categorical data
    \item Splitting dataset into training and test set 
    \item Feature scaling
    
\end{itemize}
\begin{figure}
    \centering
    \includegraphics[width=0.5\textwidth]{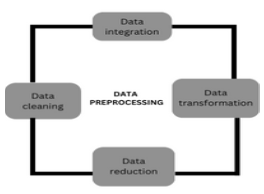}
    \caption{Data preprocessing}
    \label{fig:Dp}
\end{figure}

\subsection{Training and Testing }
The primary distinction between test data and training data is that test data is used to verify model accuracy whereas training data is a subset of the original data used to train a machine learning model. 
Typically, the training dataset is bigger than the test dataset. The ratio of 80, 20, 70:30, or 90:10 is frequently used to divide training and test datasets. Due to the fact that training data is utilised to train the model, test data is unfamiliar or invisible to the model. The dataset has been divided in half, with 80\% of the data being used for training and 20\% being utilised for testing. 

\begin{figure}
    \centering
    \includegraphics[width=0.5\textwidth]{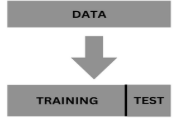}
    \caption{Train-Test Data Split}
    \label{fig:data_split}
\end{figure}

\subsection{System Configuration}
\paragraph{Hardware:}
\begin{itemize}
    \item Processor: Intel Core i5 or higher
    \item RAM: 8GB or higher
    \item Storage: Minimum of 100GB hard disk space or 256GB solid-state drive (SSD)
    \item Graphics Card: NVIDIA GeForce GTX 1050 or higher
    \item Monitor: 1920 x 1080 resolution or higher
\end{itemize}

\paragraph{Software:}
\begin{itemize}
    \item Operating System: Windows 10 or Linux Ubuntu 18.04 or later versions
    \item Python: Version 3.7 or higher
    \item Python Libraries: NumPy, Pandas, Matplotlib, Scikit-learn, TensorFlow, Keras, PyTorch, cv2, os, imutils
    \item Integrated Development Environment (IDE): PyCharm or Jupyter Notebook
    \item Web Framework: StreamLit
\end{itemize}

\subsection{Results}
\paragraph{Classification Report for Diabetes}
\begin{figure}[h]
    \centering
    \includegraphics[width=0.3\textwidth]{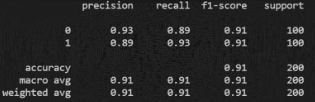} 
    \caption{Classification Report for Diabetes}
    \label{fig:diabetes_report}
\end{figure}

\paragraph{Classification Report for Heart Disease}
\begin{figure}
    \centering
    \includegraphics[width=0.5\textwidth]{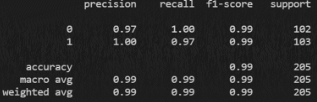}
\end{figure}

\begin{center}
    \includegraphics[width=0.3\textwidth]{10.png} 
    
\end{center}

\paragraph{Classification Report for Lung Cancer}
\begin{center}
    \includegraphics[width=0.3\textwidth]{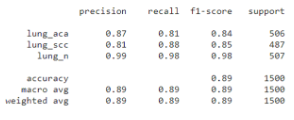} 
\end{center}

\paragraph{Classification Report for Brain Tumor}
\begin{center}
    \includegraphics[width=0.3\textwidth]{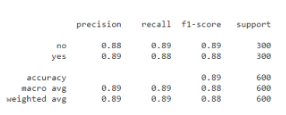} 
\end{center}

\section{Conclusion}
In conclusion, comparative analyzes have demonstrated the efficacy of ML and DL algorithms in predicting the risk of chronic diseases such as diabetes, heart disease, lung cancer and brain tumours. The DL algorithm outperformed the ML algorithm on all four datasets, achieving higher accuracy and improved clinical outcomes. The development of accurate predictive models using ML and DL techniques has important implications for the prevention, diagnosis and treatment of chronic diseases, and further research is needed to realize its full potential.

Finally, four disease accuracies were obtained:
\FloatBarrier
\begin{table}[H]
    \centering
    \begin{tabular}{|c|c|c|c|}
        \hline
        SI.No. & Diesease & Model & Accuracy \\
        \hline
        1.  & Diabetes  & Random Forest & 91.0   \\
        \hline
        2.  & Heart Disease  & Random Forest & 98.53 \\
        \hline
        3.  & Lung Cancer  & CNN(Sequential) & 89.0  \\
        \hline
        4.  & Brain Tumour  & CNN(VGG16) & 89.0   \\
        \hline
       
    \end{tabular}
    \caption{Accuracies of all four Chronic Diesease}
    \label{tab:example}
\end{table}


\end{document}